\documentclass[10pt,twocolumn,letterpaper]{article}

\usepackage{cvpr}
\usepackage{times}
\usepackage{epsfig}
\usepackage{graphicx}
\usepackage{amsmath}
\usepackage{amssymb}
\usepackage{multirow}

\usepackage{gensymb}
\usepackage{makecell}
\usepackage{hhline}

\usepackage[skip=7pt]{caption}

\newcommand{\erran}[1]{{\color{green}[Erran: #1]}}

\newcommand{\comment}[1]{}

\newcommand{\Img}{\mathbf{I}}

\newcommand{\Disp}{L}
\newcommand{\DispM}{\mathbf{L}}
\newcommand{\DepthM}{\mathbf{D}}
\newcommand{\Prob}{\mathbf{P}}
\newcommand{\Var}{\mathbf{V}}
\newcommand{\Sdv}{\mathbf{\sigma}}
\newcommand{\Interval}{\mathbf{C}}
\newcommand{\QuaterImgSize}{\frac{W}{4}\times\frac{H}{4}}
\newcommand{\HalfImgSize}{\frac{W}{2}\times\frac{H}{2}}
\newcommand{\FullImgSize}{W\times H}

\newcommand{\boldstartspace}[1]{\vspace{0.1in}\noindent\textbf{#1}}

\usepackage[pagebackref=true,breaklinks=true,letterpaper=true,colorlinks,bookmarks=false]{hyperref}

\cvprfinalcopy 

\def\cvprPaperID{2543} 

\ifcvprfinal\fi

\makeatletter
\def\blfootnote{\xdef\@thefnmark{}\@footnotetext}
\makeatother

\begin{document}

\title{Deep Stereo using Adaptive Thin Volume Representation \\with Uncertainty Awareness}

\author{Shuo Cheng$^{1}$\textsuperscript{*}\qquad 
Zexiang Xu$^{1}$\textsuperscript{*}\qquad
Shilin Zhu$^{1}$\qquad\\
Zhuwen Li$^{2}$\qquad
Li Erran Li$^{3,4}$\qquad
Ravi Ramamoorthi$^{1}$\qquad
Hao Su$^{1}$\qquad\\
$^{1}$University of California, San Diego \quad\quad $^{2}$Nuro Inc. \quad\quad
$^{3}$Columbia University \quad\quad $^{4}$Scale AI\\
{\tt\small \{scheng, shz338, haosu\}@eng.ucsd.edu \quad \{zexiangxu, ravir\}@cs.ucsd.edu }\\ 
{\tt\small \{lzhuwen, erranlli\}@gmail.com}
\vspace{-3.5mm}
}

\maketitle
\thispagestyle{empty}
\blfootnote{\textsuperscript{*} Equal contribution.}
\begin{abstract}
	\vspace{-2mm}
	We present Uncertainty-aware Cascaded Stereo Network (UCS-Net)  
	for 3D reconstruction from multiple RGB images. 
	Multi-view stereo (MVS) aims to reconstruct fine-grained scene geometry from multi-view images. 
	Previous learning-based MVS methods estimate per-view depth using plane sweep volumes with a fixed depth hypothesis at each plane;
	this generally requires densely sampled planes for desired accuracy, and it is very hard to achieve high-resolution depth.
	In contrast, we propose adaptive thin volumes (ATVs); 
	in an ATV, the depth hypothesis of each plane is spatially varying, 
	which adapts to the uncertainties of previous per-pixel depth predictions. 	 
	Our UCS-Net has three stages: the first stage processes a small standard plane sweep volume to predict low-resolution depth;
	two ATVs are then used in the following stages to refine the depth with higher resolution and higher accuracy.  
	Our ATV consists of only a small number of planes;
	yet, it efficiently partitions local depth ranges within learned small intervals. 
	In particular, we propose to use variance-based uncertainty estimates to 
	adaptively construct ATVs; 
	this differentiable process introduces reasonable and fine-grained spatial partitioning.
	Our multi-stage framework progressively sub-divides the vast scene space with increasing depth resolution and precision,
	which enables scene reconstruction with high completeness and accuracy in a coarse-to-fine fashion.
	We demonstrate that our method achieves superior performance compared with state-of-the-art benchmarks 
	on various challenging datasets.
	\vspace{-3mm}
\end{abstract}

\section{Introduction}

\begin{figure}[t]
	\centering
	\includegraphics[width=\linewidth]{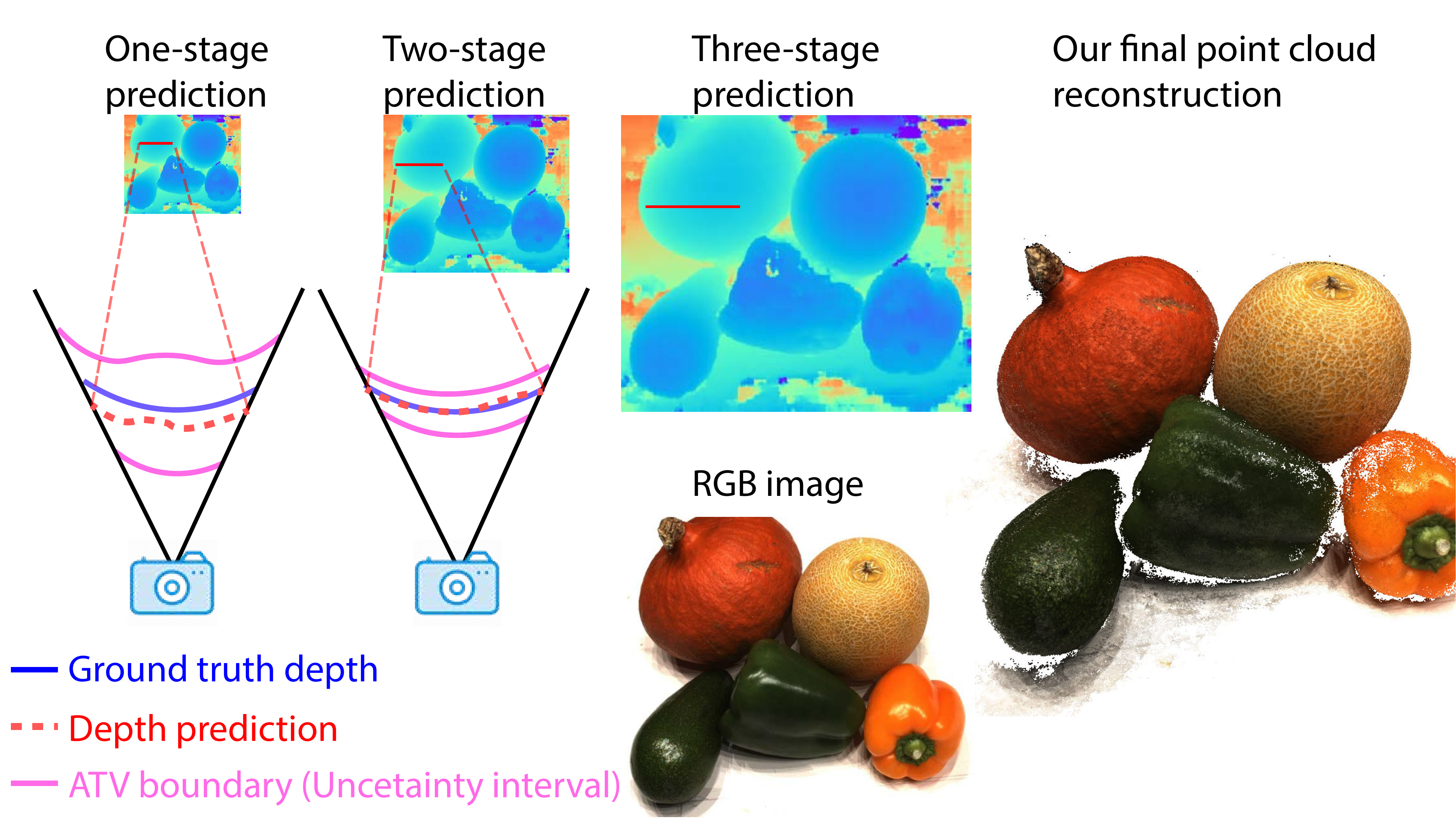}
	\caption{
		Our UCS-Net leverages adaptive thin volumes (ATVs) to progressively 
		reconstruct a highly accurate high-resolution depth map through multiple stages.
		We show the input RGB image, depth predictions with increasing sizes from three stages, and our final point cloud reconstruction 
		obtained by fusing multiple depth maps.
		We also show local 2D slices of our ATVs around a pixel (red dot).
		Note that, our ATVs become thinner after a stage because of reduced uncertainty.
		}
	\label{fig:teaser}
	\vspace{-6mm}
\end{figure}

Inferring 3D scene geometry from captured images is a core problem in computer vision and graphics 
with applications in 3D visualization, scene understanding, robotics and autonomous driving.
Multi-view stereo (MVS) aims to reconstruct dense 3D representations from multiple images with calibrated cameras.
Inspired by the success of deep convolutional neural networks (CNN), 
several learning-based MVS methods have been presented \cite{ji2017surfacenet, kar2017learning, wu2017marrnet,huang2018deepmvs,tang2018ba}; 
the most recent work leverages cost volumes in a learning pipeline \cite{yao2018mvsnet,im2018dpsnet}, 
and outperforms many traditional MVS methods \cite{furukawa2010accurate}.

At the core of the recent success on MVS \cite{yao2018mvsnet,im2018dpsnet} is the application of 3D CNNs on plane sweep cost volumes
to effectively infer multi-view correspondence.
However, such 3D CNNs involve massive memory usage for depth estimation with high accuracy and completeness.
In particular, for a large scene, high accuracy requires sampling a large number of sweeping planes and 
high completeness requires reconstructing high-resolution depth maps.
In general, given limited memory, there is an undesired trade-off between accuracy (more planes) and completeness (more pixels) in previous work \cite{yao2018mvsnet,im2018dpsnet}.

\comment{
For example, to ensure enough accuracy, MVSNet~\cite{yao2018mvsnet} leverages a dense set of sweeping planes -- 256 planes; 
however, the network can only reconstruct depth maps with a resolution that is just $1/16$ to the original image resolution (i.e. $1/4$ to each dimension),
which limits the completeness of the reconstruction.
In general, it is highly challenging to reconstruct high-resolution depth without losing accuracy.    
}

Our goal is to achieve \textit{highly accurate and highly complete reconstruction} with \textit{low memory and computation consumption} at the same time. 
To do so, we propose a novel learning-based uncertainty-aware multi-view stereo framework, 
which utilizes multiple small volumes, instead of a large standard plane sweep volume, 
to progressively regress high-quality depth in a coarse-to-fine fashion.
A key in our method is that 
we propose novel adaptive thin volumes (ATVs, see Fig.~\ref{fig:teaser}) to achieve efficient spatial partitioning.

\begin{figure*}[t]
	\centering
	\includegraphics[width=\linewidth]{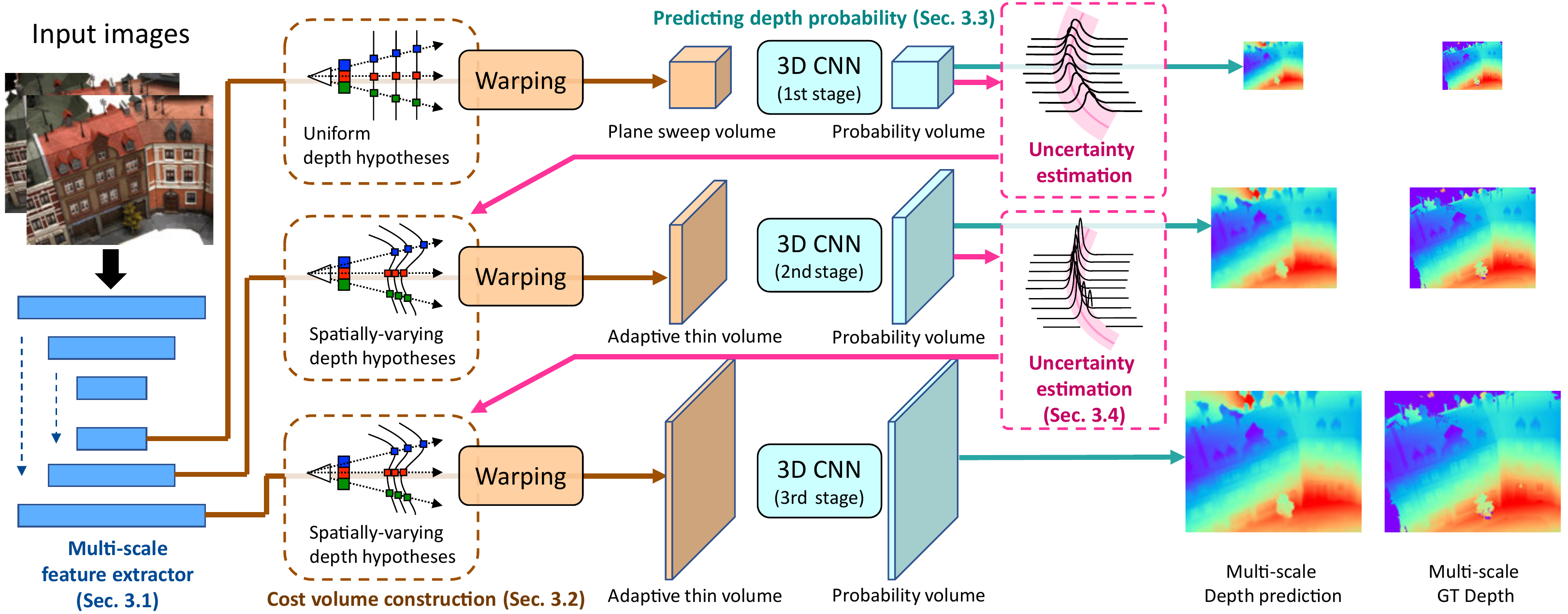}
	\caption{
		Overview of our UCS-Net. 
		Our UCS-Net leverages multi-scale cost volumes to achieve coarse-to-fine depth prediction with three cascade stages.
		The cost volumes are constructed using multi-scale deep image features from a multi-scale feature extractor.
		The last two stages utilize the uncertainty of the previous depth prediction to build adaptive thin volumes (ATVs)
		for depth reconstruction at a finer scale. We mark different parts of the network in different colors. 
		Please refer to Sec~\ref{sec:method} and the corresponding subsections for more details. 
	}
	\label{fig:ucnet}
\end{figure*}

Specifically, we propose a novel cascaded network with three stages (see Fig.~\ref{fig:ucnet}): 
\comment{
\erran{different resolution means different depth resolution here? need to be explicit since we have image resolution as well}
}
each stage of the cascade predicts a depth map with a different size; 
each following stage constructs an ATV to refine the predicted depth from the previous stage with higher pixel resolution and finer depth partitioning.
The first stage uses a small standard plane sweep volume with low image resolution and relatively sparse depth planes 
-- 64 planes that are fewer than the number of planes (256 or 512) in previous work~\cite{yao2018mvsnet,yao2019recurrent};
the following two stages use ATVs with higher image resolutions and significantly fewer depth planes -- only 32 and 8 planes.
While consisting of a very small number of planes, our ATVs are constructed within \emph{learned local depth ranges}, 
which enables \emph{efficient and fine-grained spatial partitioning} for accurate and complete depth reconstruction. 

This is made possible by the novel uncertainty-aware construction of an ATV.  
In particular, we leverage the variances of the predicted per-pixel depth probabilities, 
and infer the uncertainty intervals (as shown in Fig.~\ref{fig:teaser}) 
by calculating variance-based confidence intervals of the per-pixel probability distributions for the ATV construction. 
Specifically, we apply the previously predicted depth map as a central curved plane, 
and construct an ATV around the central plane within local per-pixel uncertainty intervals.
In this way, we explicitly express the uncertainty of the depth prediction at one stage, 
and embed this knowledge into the input volume for the next stage.

Our variance-based uncertainty estimation is differentiable and 
we train our UCSNet from end to end with depth supervision for the predicted depths from all three stages.
Our network can thus learn to optimize the estimated uncertainty intervals,
to make sure that an ATV is constructed with proper depth coverage that is both large enough -- to try to cover ground truth depth -- and small enough 
 -- to enable accurate reconstruction for the following stages. 
Overall, our multi-stage framework can
progressively sub-divide the local space at a finer scale in a reasonable way, which leads to high-quality depth reconstruction.
We demonstrate that our novel UCS-Net outperforms the state-of-the-art learning-based MVS methods on various datasets.

\comment{Our method is able 
Our framework is computation- and memory- efficient, in which every single stage deals with a small volume with either small image-wise resolution or small depth-wise resolution.
More importantly, our coarse-to-fine framework enables the following stages to 
progressively sub-divide the local space at a finer scale and achieve better depth reconstruction.}

\section{Related Work}

Multi-view stereo is a long-studied vision problem with many traditional approaches 
\cite{seitz2006comparison,pons2003variational,kutulakos2000theory,kolmogorov2002multi,kang2001handling,esteban2004silhouette,de1999poxels,furukawa2010accurate,schonberger2016pixelwise}.
Our learning-based framework leverages the novel spatial representation, ATV to reconstruct high-quality depth for fine-grain scene reconstruction. 
In this work, we mainly discuss spatial representation for 3D reconstruction and deep learning based multi-view stereo. 

\noindent\textbf{Spatial Representation for 3D Reconstruction. }
Existing methods can be categorized based on learned 3D representations. 
Volumetric based approaches partition the space into a regular 3D volume with millions of small voxels \cite{ji2017surfacenet, kar2017learning, wu2017marrnet, wu2018learning, zhang2018learning, richter2018matryoshka}, and the network predicts if a voxel is on the surface or not. 
Ray tracing can be incorporated into this voxelized structure  \cite{tulsiani2017multi, paschalidou2018raynet, ulusoy2015towards}. 
The main drawback of these methods is computation and memory inefficiency, given that most voxels are not on the surface. 
Researchers have also tried to reconstruct point clouds \cite{insafutdinov2018unsupervised, furukawa2010accurate, lhuillier2005quasi, wang2018mvpnet, lin2018learning, achlioptas2018learning}, however the high dimensionality of a point cloud often results in noisy outliers since a point cloud does not efficiently encode connectivity between points. 
Some recent works utilize single or multiple images to reconstruct a point cloud given strong shape priors \cite{fan2017point, insafutdinov2018unsupervised, lin2018learning}, which cannot be directly extended to large-scale scene reconstruction. 
Recent work also tried to directly reconstruct surface meshes \cite{ladicky2017point, kanazawa2018learning, wang2018pixel2mesh, henderson2019learning, sinha2017surfnet, kato2018neural}, deformable shapes \cite{kanazawa2018end, kanazawa2018learning}, and some learned implicit distance functions \cite{dai2017shape, riegler2017octnetfusion, mescheder2018occupancy, chen2018learning}. These reconstructed surfaces often look smoother than point-cloud-based approaches, but often lack high-frequency details. 
A depth map represents dense 3D information that is perfectly aligned with a reference view; depth reconstruction has been demonstrated in many previous works on reconstruction with both single view \cite{eigen2015predicting, ummenhofer2017demon, garg2016unsupervised, godard2017unsupervised, zhou2017unsupervised} and multiple views \cite{campbell2008using, tola2012efficient, hartmann2017learned, galliani2015massively, schonberger2016pixelwise, yao2017relative, schonberger2016pixelwise}. 
Some of them leverage normal information as well \cite{galliani2015massively, galliani2016just}.  
\comment{Despite this advantage, depth value correlates with camera extrinsic parameters, making accurate estimation harder to achieve.
}
In this paper, we present ATV, a novel spatial representation for depth estimation; we use two ATVs to progressively partition local space, which is the key to achieve coarse-to-fine reconstruction.

\noindent\textbf{Deep Multi-View Stereo (MVS). } The traditional MVS pipeline mainly relies on photo-consistency constraints to infer the underlying 3D geometry, but usually performs poorly on texture-less or occluded areas, or under complex lighting environments. 
To overcome such limitations, many deep learning-based  MVS methods have emerged in the last two years, including regression-based approaches \cite{yao2018mvsnet,im2018dpsnet}, classification-based approaches \cite{huang2018deepmvs} and approaches based on recurrent- or iterative- style architectures
\cite{yao2019recurrent,zhou2018deeptam,chen2019point} and many other approaches \cite{kendall2017end, paschalidou2018raynet, batsos2018cbmv, shin20193d}. 
Most of these methods build a single cost volume with uniformly sampled depth hypotheses by projecting 2D image features into 3D space, and then use a stack of either 2D or 3D CNNs to infer the final depth \cite{yao2018mvsnet,flynn2016deepstereo,xu2019deep}. 
However, a single cost volume often requires a large number of depth planes to achieve enough reconstruction accuracy, and it is difficult to reconstruct high-resolution depth, limited by the memory bottleneck.
R-MVSNet~\cite{yao2019recurrent} leverages recurrent networks to sequentially build a cost volume with a high depth-wise sampling rate (512 planes). 
In contrast, we apply an adaptive sampling strategy with ATVs, which enables more efficient spatial partitioning with a higher depth-wise sampling rate using fewer depth planes (104 planes in total, see Tab.~\ref{tab:uncertainty}), and our method achieves significantly better reconstruction than R-MVSNet (see Tab.~\ref{tab:dtu} and Tab.~\ref{tab:tat}).
On the other hand, Point-MVSNet~\cite{chen2019point} densifies a coarse reconstruction within a predefined local spatial range for better reconstruction with learning-based refinement.
We propose to refine depth in a learned local space with adaptive thin volumes to obtain accurate high-resolution depth, which leads to better reconstruction than Point-MVSNet and other state-of-the-art methods (see Tab.~\ref{tab:dtu} and Tab.~\ref{tab:tat}).

\section{Method}

\label{sec:method}
\comment{All previous works \cite{huang2018deepmvs,yao2018mvsnet,im2018dpsnet} apply deep CNNs on one single cost volume; 
in such volumes, the number of sweeping planes expresses the depth resolution and determines the precision of depth estimation, 
which is, however, strictly limited by system memory resources.
We propose a novel coarse-to-fine deep stereo approach; 
we leverage multiple small volumes that progressively sub-partitions local space around the actual depth at a finer scale, and achieve highly accurate depth reconstruction.
While each individual volume merely consists of a small number (16 or 32) of sweeping planes, the overall depth resolution increases exponentially in the process, and thus allows higher accuracy as well as less 
computation-  and memory- efficiency complexity.
}
 
Some recent works aim to improve learning-based MVS methods. 
Recurrent networks \cite{yao2019recurrent} have been utilized to achieve fine depth-wise partitioning for high accuracy; 
a PointNet-based method \cite{chen2019point} is also presented to densify the reconstruction for high completeness.
Our goal is to reconstruct high-quality 3D geometry with both high accuracy and high completeness.
To this end, we propose a novel uncertainty-aware cascaded network (UCS-Net) to reconstruct highly accurate per-view depth with high resolution. 

Given a reference image $\Img_1$ and $N-1$ source images $\{\Img_i\}_{i=2}^N$, 
our UCS-Net progressively regresses a fine-grained depth map at the same resolution as the reference image.
We show the architecture of the UCS-Net in Fig. \ref{fig:ucnet}.
Our UCS-Net first leverages a 2D CNN to extract multi-scale deep image features at three resolutions (Sec.~\ref{sec:feature}).
Our depth prediction is achieved through three stages, 
which leverage multi-scale image features to predict multi-resolution depth maps.
In these stages, we construct multi-scale cost volumes (Sec.~\ref{sec:volume}), where each volume is a plane sweep volume or an adaptive thin volume (ATV).
We then apply 3D CNNs to process the cost volumes to predict per-pixel depth probability distributions, 
and a depth map is reconstructed from the expectations of the distributions (Sec.~\ref{sec:depth}).
To achieve efficient spatial partitioning, 
we utilize the uncertainty of the depth prediction to construct ATVs as cost volumes for the last two stages (Sec.~\ref{sec:uncertainty}).
Our multi-stage network effectively reconstructs depth in a coarse-to-fine fashion (Sec.~\ref{sec:coarse2fine}).

\subsection{Multi-scale feature extractor}
\label{sec:feature}
Previous methods use downsampling layers \cite{yao2018mvsnet,yao2019recurrent} or a UNet \cite{xu2019deep} to extract deep features 
and build a plane sweep volume at a single resolution.
To reconstruct high-resolution depth, we introduce a multi-scale feature extractor, 
which enables constructing multiple cost volumes at different scales for multi-resolution depth prediction.
As schematically shown in Fig.~\ref{fig:ucnet}, 
our feature extractor is a small 2D UNet \cite{ronneberger2015u}, which has an encoder and a decoder with skip connections.
The encoder consists of a set of convolutional layers followed by BN (batch normalization) and ReLu activation layers; 
we use stride = 2 convolutions to downsample the original image size twice.
The decoder upsamples the feature maps, convolves the upsampled features and the concatenated features from skip links,
 and also applies BN and Relu layers.
Given each input image $\Img_i$, the feature extractor provides three scale feature maps, $F_{i,1}$, $F_{i,2}$, $F_{i,3}$, from the decoder for the following cost volume construction.
We represent the original image size as $\FullImgSize$, where $W$ and $H$ denote the image width and height;
correspondingly, $F_{i,1}$, $F_{i,2}$ and $F_{i,3}$ have resolutions of $\QuaterImgSize$, $\HalfImgSize$ and $\FullImgSize$, and their numbers of channels are 32, 16 and 8 respectively.
Please refer to Tab.~\ref{fig:2d_unet} in the appendix for the details of our 2D CNN architecture.
Our multi-scale feature extractor allows for the high-resolution features to properly incorporate the information at lower resolutions through the learned upsampling process;
thus in the multi-stage depth prediction, each stage is aware of the meaningful feature knowledge used in previous stages, 
which leads to reasonable high-frequency feature extraction.

\subsection{Cost volume construction}
\label{sec:volume}
We construct multiple cost volumes at multiple scales by warping the extracted feature maps, $F_{i,1}$, $F_{i,2}$, $F_{i,3}$ from source views to a reference view.
Similar to previous work, this process is achieved through differentiable unprojection and projection. 
In particular, given camera intrinsic and extrinsic matrices $\{K_i, T_i\}$ for each view $i$, 
the $4\times 4$ warping matrix at depth $d$ at the reference view is given by:
\begin{align}
	H_i(d) = K_i T_i T^{-1}_1K^{-1}_1.
	\label{eqn:warp}
\end{align}
In particular, when warping to a pixel in the reference image $\Img_1$ at location $(x,y)$ and depth $d$, 
$H_i(d)$ multiplies the homogeneous vector $(xd,yd,d,1)$ to finds its corresponding pixel location in each $\Img_i$ in homogeneous coordinates.

Each cost volume consists of multiple planes; 
we use $\DispM_{k,j}$ to denote the depth hypothesis of the $j$th plane at the $k$th stage, 
and $\DispM_{k,j}(x)$ represents its value at pixel $x$.
At stage $k$, 
once we warp per-view feature maps $F_{i,k}$ at all depth planes with corresponding hypotheses $\DispM_{k,j}$, 
we calculate the variance of the warped feature maps across views at each plane to construct a cost volume.
We use $D_k$ to represent the number of planes for stage $k$.
For the first stage, we build a standard plane sweep volume, 
whose depth hypotheses are of constant values, i.e. $\DispM_{1,j}(x) = d_j$.
We uniformly sample $\{d_j\}_{j=1}^{D_1}$ from a pre-defined depth interval $[d_{min},d_{max}]$ to 
construct the volume, in which each plane is constructed using $H_i(d_j)$ to warp multi-view images.
For the second and third stages, we build novel adaptive thin volumes, 
whose depth hypotheses have spatially-varying depth values according to pixel-wise uncertainty estimates of the previous depth prediction.
In this case, we calculate per-pixel per-plane warping matrices by setting $d=\DispM_{k,j}(x)$ in Eqn.~\ref{eqn:warp} to warp images and construct cost volumes.
Please refer to Sec.~\ref{sec:uncertainty} for uncertainty estimation.

\subsection{Depth prediction and probability distribution}
\label{sec:depth}
At each stage, we apply a 3D CNN to process the cost volume, infer multi-view correspondence and predict depth probability distributions.
In particular, we use a 3D UNet similar to \cite{yao2018mvsnet}, 
which has multiple downsampling and upsampling 3D convolutional layers
to reason about scene geometry at multiple scales.
We apply depth-wise softmax at the end of the 3D CNNs to predict per-pixel depth probabilities.
Our three stages use the same network architecture without sharing weights, 
so that each stage learns to process its information at a different scale.
Please refer to Tab.~\ref{fig:3d_unet} in the appendix for the details of our 3D CNN architecture.

The 3D CNN at each stage predicts a depth probability volume 
that consists of $D_k$ depth probability maps $\Prob_{k,j}$ associated with the depth hypotheses $\DispM_{k,j}$.
$\Prob_{k,j}$ expresses per-pixel depth probability distributions, 
where $\Prob_{k,j}(x)$ represents how probable the depth at pixel $x$ is $\DispM_{k,j}(x)$. 
A depth map $\hat{\DispM}_k$ at stage $k$ is reconstructed by weighted sum:
\begin{equation}
 \hat{\DispM}_k(x) = \sum_{j=1}^{D_k}{\DispM_{k,j}(x) \cdot \Prob_{k,j}(x)}.
 \label{eqn:expect}
\end{equation}

\subsection{Uncertainty estimation and ATV}
\label{sec:uncertainty}
The key for our framework is to progressively sub-partition the local space and refine the depth prediction with increasing resolution and accuracy.
To do so, we construct novel ATVs for the last two stages, which have curved sweeping planes with spatially-varying depth hypotheses 
(as illustrated in Fig.~\ref{fig:teaser} and Fig.~\ref{fig:ucnet}),
based on uncertainty inference of the predicted depth in its previous stage.

\comment{
\begin{figure}[t]
    \centering
    \includegraphics[width=\linewidth]{images/iccv_method.pdf}
    \caption{At a cascade stage, we predict a depth map (middle) from input RGB images (left), and infer the uncertainty of the prediction, expressed by a confidence interval. 
    On the right, we show the predicted depth probabilities (connected blue dots) of a pixel (green point), depth prediction (red dash line), the ground truth depth (blue dash line) and confidence intervals of $\sigma$, $1.5\sigma$ and $2\sigma$.}
    \label{fig:method}
    \vspace{-3mm}
\end{figure}
}

Given a set of depth probability maps, 
previous work only utilizes the expectation of the per-pixel distributions 
(using Eqn.~\eqref{eqn:expect}) to determine an estimated depth map.
For the first time, we leverage the variance of the distribution for uncertainty estimation, 
and construct ATVs using the uncertainty.
In particular, the variance $\hat{\Var}_{k}(x)$ of the probability distribution at pixel $x$ and stage $k$ is calculated as:
\begin{gather}
\hat{\Var}_{k}(x) = \sum_{j=1}^{D_k}\Prob_{k,j}(x) \cdot(\DispM_{k,j}(x)-\hat{\DispM}_{k}(x))^2,
\end{gather}
and the corresponding standard deviation is $\hat{\Sdv}_k(x) = \sqrt{\hat{\Var}_k}$.
Given the depth prediction $\hat{\DispM}_{k}(x)$ and its variance $\hat{\Sdv}_k(x)^2$ at pixel $x$, 
we propose to use a variance-based confidence interval to measure the uncertainty of the prediction:
\begin{equation}
	\Interval_{k}(x) = [\hat{\DispM}_{k}(x) - \lambda\hat{\Sdv}_k(x), \hat{\DispM}_{k}(x)+\lambda\hat{\Sdv}_k(x)],
	\label{eqn:interval}
\end{equation}
where $\lambda$ is a scalar parameter that determines how large the confidence interval is.
For each pixel $x$, we uniformly sample $D_{k+1}$ depth values from $\Interval_{k}(x)$ of the $k$th stage, 
to get its depth values $\DispM_{k+1,1}(x)$, $\DispM_{k+1,2}(x)$,...,$\DispM_{k+1,D_{k+1}}(x)$ of the depth planes for stage $(k+1)$.
In this way, we construct $D_{k+1}$ spatially-varying depth hypotheses $\DispM_{k+1,j}$, which 
form the ATV for stage $(k+1)$.

\begin{figure}[t]
    \centering
    \includegraphics[width=\linewidth]{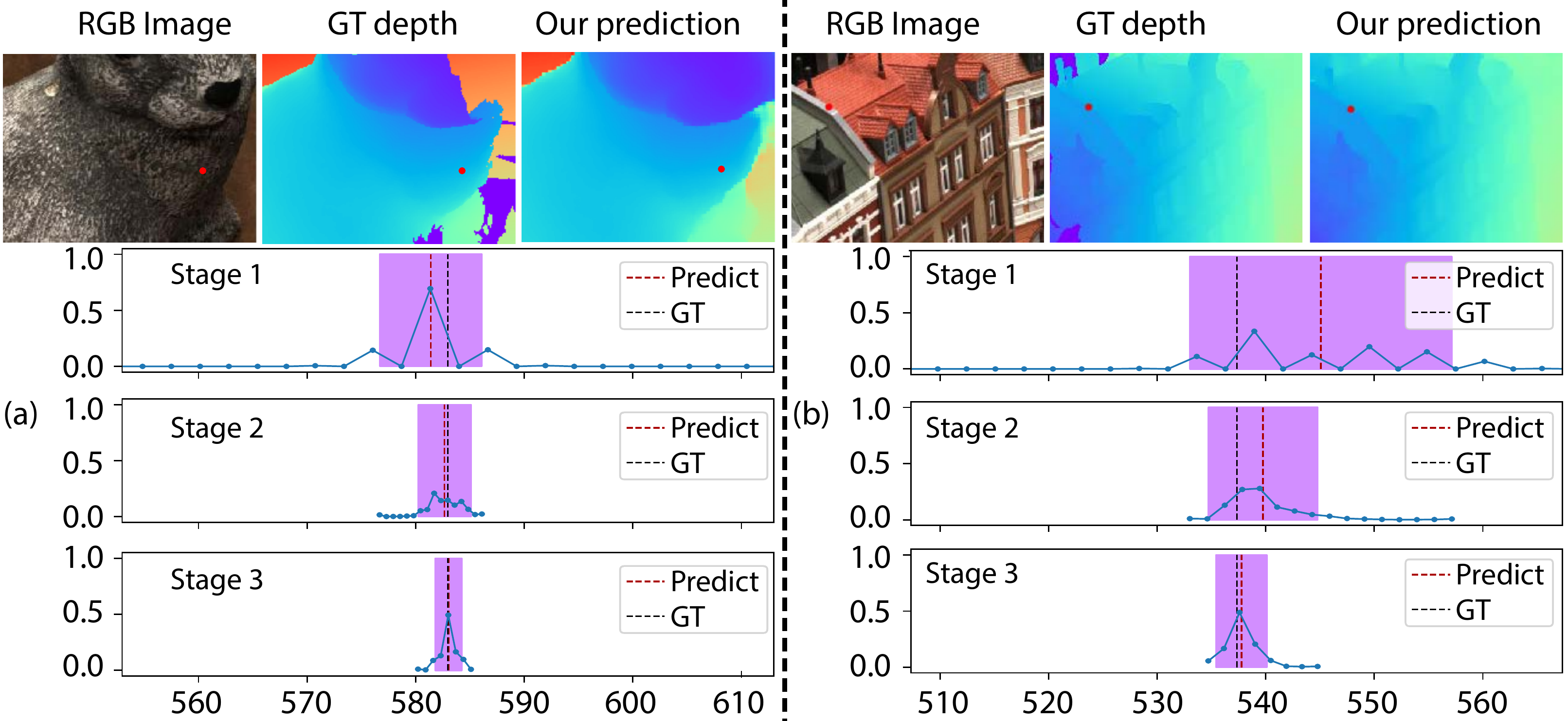}
	\caption{ 
	We illustrate detailed depth and uncertainty estimation of two examples.
	On the top, we show the RGB image crops, predicted depth and ground truth depth.
	On the bottom, 
	we show the details of of two pixels (red points in the images) with predicted depth probabilities (connected blue dots) , 
	depth prediction (red dash line), the ground truth depth (black dash line) 
	and uncertainty intervals (purple) in the three stages.}
    \label{fig:method}
\end{figure}

The estimated $\Interval_{k}(x)$ expresses the uncertainty interval of the prediction $\hat{\DispM}_{k}(x)$,
which determines the physical thickness of an ATV at each pixel.
In Fig.~\ref{fig:method}, 
we show two actual examples with two pixels and their estimated uncertainty intervals $\Interval_{k}(x)$ around the predictions (red dash line).
The $\Interval_{k}$ essentially depicts a probabilistic local space around the ground truth surface, 
and the ground truth depth is located in the uncertainty interval with a very high confidence. 
Note that, our variance-based uncertainty estimation is differentiable, 
which enables our UCS-Net to learn to adjust the probability prediction at each stage to 
achieve optimized intervals and corresponding ATVs for following stages in an end-to-end training process.
As a result, the spatially varying depth hypotheses in ATVs naturally adapt to the uncertainty of depth predictions, 
which leads to highly efficient spatial partitioning.

\subsection{Coarse-to-fine prediction}
\label{sec:coarse2fine}
Our UCS-Net leverages three stages to reconstruct depth at multiple scales from coarse to fine, which generally supports different numbers ($D_k$) of planes in each stage.
In practice, we use $D_1=64$, $D_2=32$ and $D_3=8$ to construct a plane sweep volume and two ATVs 
with sizes of $\QuaterImgSize \times 64$, $\HalfImgSize \times 32$ and $H\times W\times 8$ to estimate depth at corresponding resolutions.
While our two ATVs have small numbers ($32$ and $8$) of depth planes,
they in fact partition local depth ranges at finer scales than the first stage volume;
this is achieved by our novel uncertainty-aware volume construction process which adaptively controls local depth intervals.
This efficient usage of a small number of depth planes enables the last two stages to deal with higher pixel-wise resolutions given the limited memory,
which makes fine-grained depth reconstruction possible.
Our novel ATV effectively expresses the locality and uncertainty in the depth prediction, 
which enables state-of-the-art depth reconstruction results with high accuracy and high completeness through a coarse-to-fine framework.

\subsection{Training details}
\label{sec:details}
\noindent\textbf{Training set.} 
We train our network on the DTU dataset~\cite{aanaes2016large}. 
We split the dataset into training, validate and testing set, and create ground truth depth similar to \cite{yao2018mvsnet}.
In particular, we apply Poisson reconstruction \cite{kazhdan2013screened} on the point clouds in DTU, and render the surface at the captured views with three
resolutions, $\QuaterImgSize$, $\HalfImgSize$ and the original $W\times H$.
In particular, we use $W\times H = 640\times 512$ for training.

\noindent\textbf{Loss function.}
Our UCS-Net predicts depth at three resolutions; 
we apply $L1$ loss on depth prediction at each resolution with the rendered ground truth at the same resolution.
Our final loss is the combination of the three $L1$ losses.

\noindent\textbf{Training policy.}
We train our full three-stage network from end to end for 60 epochs.
We use Adam optimizer with an initial learning rate of $0.0016$.
We use 8 NVIDIA GTX 1080Ti GPUs to train the network with a batch size of 16 (mini-batch size of 2 per GPU).

\section{Experiments}
We now evaluate our UCS-Net.
We do benchmarking on the DTU and Tanks and Temple datasets.
We then justify the effectiveness of the designs of our network, 
in terms of uncertainty estimation and multi-stage prediction. 

\begin{table}[t!]
	\centering
	\begin{tabular}[width=\linewidth]{r|p{1.4cm}p{1.4cm}p{1.4cm}}
		\Xhline{2pt}
		Method & Acc. & Comp. & Overall \\ 
		\Xhline{2pt}
		Camp \cite{campbell2008using} & 0.835 & 0.554 & 0.695\\
		
		Furu \cite{furukawa2010accurate} & 0.613 & 0.941 & 0.777\\
		
		Tola \cite{tola2012efficient} & 0.342 & 1.190 & 0.766\\
		
		Gipuma \cite{galliani2015massively} & \textbf{0.283} & 0.873 & 0.578\\
		
        SurfaceNet \cite{sinha2017surfnet} & 0.450 & 1.040 & 0.745\\
        
        MVSNet \cite{yao2018mvsnet} & 0.396 & 0.527 & 0.462\\

        R-MVSNet \cite{yao2019recurrent} & 0.383 & 0.452 & 0.417\\

        Point-MVSNet \cite{chen2019point} & 0.342 & 0.411 & \underline{0.376}\\
        \Xhline{2pt}
        Our 1st stage &0.548 & 0.529 & 0.539\\

        Our 2nd stage & 0.401 & \underline{0.397} & 0.399\\

        Our full model & \underline{0.338} & \textbf{0.349} & \textbf{0.344}\\

		\Xhline{2pt}
	\end{tabular}
	\caption{Quantitative results of accuracy, completeness and overall on the DTU testing set. Numbers represent distances in millimeters and smaller means better.}
	\vspace{-0.12in}
	\label{tab:dtu}
\end{table}

\begin{table*}[th!]
	\centering
	\begin{tabular}[width=\linewidth]{r|p{1.1cm}p{1.1cm}p{1.1cm}p{1.1cm}p{1.24cm}p{1.1cm}p{1.1cm}p{1.24cm}p{1.1cm}}
		\Xhline{2pt}
		 Method &  Mean & Family & Francis & Horse & Lighthouse & M60 & Panther & Playground & Train\\ 
		\Xhline{2pt}
		 MVSNet\cite{yao2018mvsnet}  & 43.48 & 55.99 & 28.55 & 25.07 & \underline{50.79} & \underline{53.96} & \underline{50.86} & 47.90 & 34.69 \\
		 R-MVSNet\cite{yao2019recurrent}   & 48.40 & 69.96 & 46.65 & 32.59 & 42.95 & 51.88 & 48.80 & 52.00 & 42.38 \\
		 Dense-R-MVSNet\cite{yao2019recurrent} & \underline{50.55} &  \underline{73.01} &  \textbf{54.46}
		 &  \textbf{43.42} & 43.88 & 46.80 & 46.69 & 50.87 & \underline{45.25} \\
		 Point-MVSNet\cite{chen2019point}   & 48.27 & 61.79 & 41.15 & 34.20 & \underline{50.79} & 51.97 & 50.85 & \underline{52.38} & 43.06 \\
        \Xhline{2pt}
		Our full model                     & \textbf{54.83} & \textbf{76.09} & \underline{53.16} & \underline{43.03} &  \textbf{54.00} &  \textbf{55.60} & \textbf{51.49} & \textbf{57.38} &  \textbf{47.89} \\
		\Xhline{2pt}
	\end{tabular}
	\vspace{+1mm}
	\caption{Quantitative results of F-scores (higher means better) on Tanks and Temples.}
	\vspace{+1mm}
	\label{tab:tat}
\end{table*}

\boldstartspace{Evaluation on the DTU dataset \cite{aanaes2016large}.}
We evaluate our method on the DTU testing set. 
To reconstruct the final point cloud, we follow \cite{galliani2015massively} to fuse the depth from multiple views;
we use this fusion method for all our experiments.
For fair comparisons, we use the same view selection, image size and initial depth range as in \cite{yao2018mvsnet} 
with $N=5$, $W=1600$, $H=1184$, $d_{\text{min}}=425mm$ and $d_{\text{max}}=933.8mm$;
similar settings are also used in other learning-based MVS methods \cite{chen2019point,yao2019recurrent}.
We use a NVIDIA GTX 1080 Ti GPU to run the evaluation.

\begin{figure}[t]
    \centering
    \includegraphics[width=\linewidth]{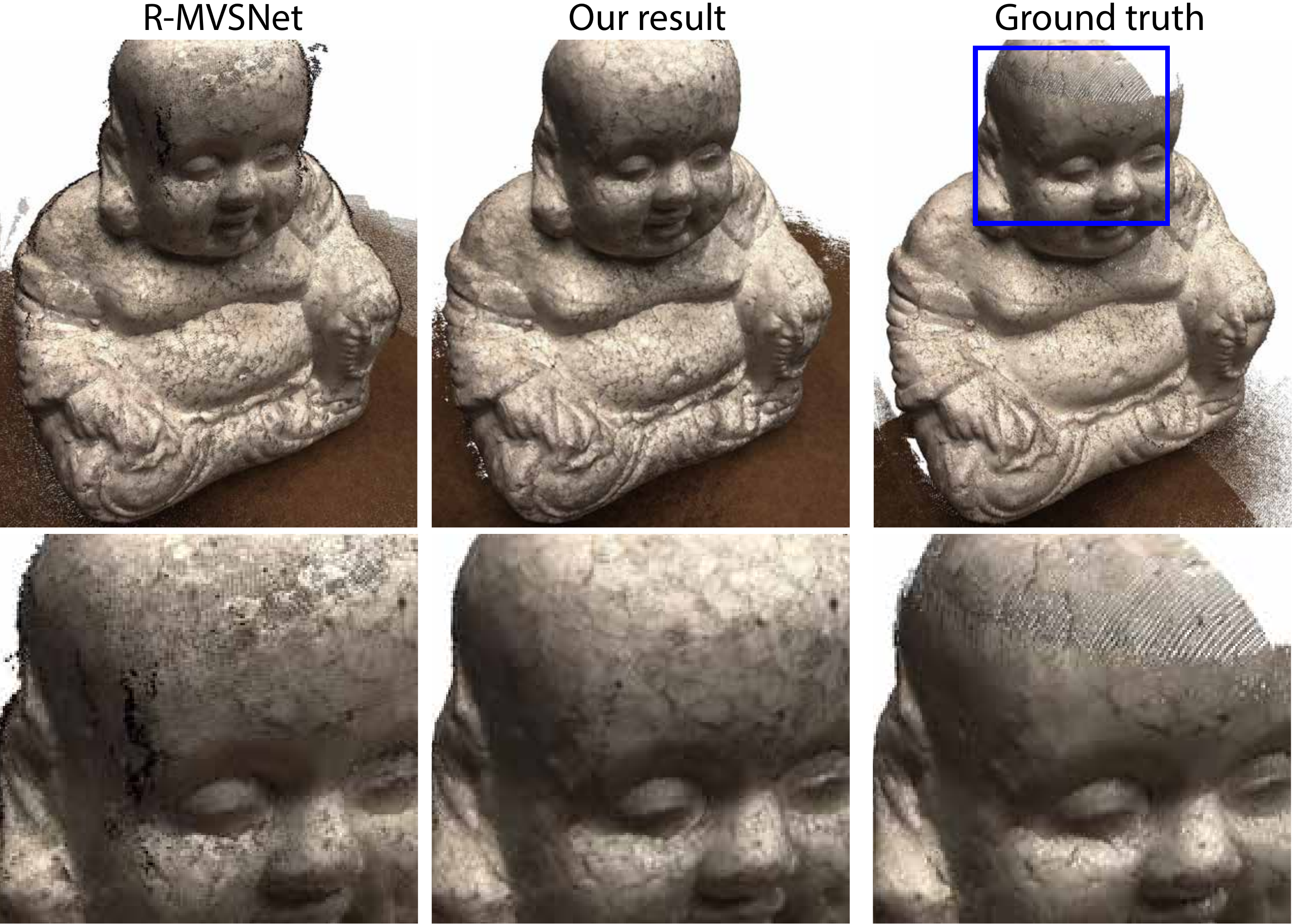}
	\caption{Comparisons with R-MVSNet on an example in the DTU dataset. We show rendered images of the point clouds of our method, R-MVSNet and the ground truth. In this example, the ground truth from scanning is incomplete. 
	We also show insets for detailed comparisons marked as a blue box in the ground truth. 
	Note that our result is smoother and has fewer outliers than R-MVSNet's result.
	}
    \label{fig:rmvs}
\end{figure}

We compare the accuracy and the completeness of the final reconstructions using the distance metric in \cite{aanaes2016large}.
We compare against both traditional methods and learning-based methods, and the average quantitative results are shown in Tab.~\ref{tab:dtu}.
While Gipuma~\cite{galliani2015massively} (a traditional method) achieves the best accuracy among all methods, 
our method has significantly better completeness and overall scores.  
Besides, our method outperforms all state-of-the-art baseline methods in terms of both accuracy and completeness.
Note that with the same input, MVSNet and R-MVSNet predict depth maps with a size of only $\QuaterImgSize$;
our final depth maps are estimated at the original image size, which are of much higher resolution 
and lead to significantly better completeness.
Meanwhile, such high completeness is obtained without losing accuracy; 
our accuracy is also significantly better thanks to our uncertainty-aware progressive reconstruction.
Point-MVSNet~\cite{chen2019point} densifies low-resolution depth within a predefined local depth range, 
which also reconstructs depth at the original image resolution; 
in contrast, our UCS-Net leverages learned adaptive local depth ranges and achieves better accuracy and completeness.  

We also show results from our intermediate low-resolution depth of the first and the second stages in Tab.~\ref{tab:dtu}.  
Note that, because of sparser depth planes, 
our first-stage results (64 planes) are worse than MVSNet (256 planes) and R-MVSNet (512 planes) that reconstruct depth at the same low resolution.
Nevertheless, our novel uncertainty-aware network introduces highly efficient spatial partitioning with ATVs in the following stages,
so that our intermediate second-stage reconstruction is already much better than the two previous methods, 
and our third stage further improves the quality and achieves the best reconstruction.

We show qualitative comparisons between our method and R-MVSNet~\cite{yao2019recurrent} in Fig.~\ref{fig:rmvs},
in which we use the released point cloud reconstruction on R-MVSNet's website for the comparison.
While both methods achieve comparable completeness in this example, 
it is very hard for R-MVSNet to achieve high accuracy at the same time, 
which introduces obvious outliers and noise on the surface.
In contrast, our method is able to obtain high completeness and high accuracy simultaneously 
as reflected by the smooth complete geometry in the image.

\boldstartspace{Evaluation on Tanks and Temple dataset \cite{knapitsch2017tanks}.}
We now evaluate the generalization of our model by testing our network trained with the DTU dataset on complex outdoor scenes in the Tanks and Temple intermediate dataset.
We use $N=5$ and $W\times H=1920\times 1056$ for this experiment.
Our method outperforms most published methods, and to the best of our knowledge, when comparing with all published learning-based methods, we achieve the best average F-score (54.83) as shown in Tab.~\ref{tab:tat}.
In particular, our method obtains higher F-scores 
than MVSNet~\cite{yao2018mvsnet} and Point-MVSNet~\cite{chen2019point} in all nine testing scenes.
Dense-R-MVSNet leverages a well-designed post-processing method and achieves slightly better performance than ours on two of the scenes,
whereas our work is focused on high-quality per-view depth reconstruction, and we use a traditional fusion technique for post-processing.
Nonetheless, thanks to our high-quality depth, our method still outperforms Dense-R-MVSNet on most of the testing scenes and achieves the best overall performance.

\boldstartspace{Evaluation of uncertainty estimation.}
One key design of our UCS-Net is leveraging differentiable uncertainty estimation for the ATV construction.
We now evaluate our uncertainty estimation on the DTU validate set.
In Tab.~\ref{tab:uncertainty}, we show the average length of our estimated uncertainty intervals, 
the corresponding average sampling distances between planes, 
and the ratio of the pixels whose estimated uncertainty intervals cover the ground truth depth in the ATVs; 
we also show the corresponding values of the standard plane sweep volume (PSV) used in the first stage, 
which has an interval length of $d_{\text{max}}-d_{\text{min}}=508.8mm$ and covers the ground truth depth with certainty.

\begin{table}[t!]
	\centering
	\begin{tabular}[width=\linewidth]{r|p{1.2cm}p{1.2cm}p{1.2cm}p{1.2cm}}
		\Xhline{2pt}
				  & Ratio    & Interval & \makecell{$D_k$} & Unit\\
		\Xhline{2pt}
		  PSV     & 100$\%$  & 508.8mm   & \makecell{64} & 7.95mm \\
        1st ATV   & 94.72$\%$ & 13.88mm & \makecell{32} & 0.43mm\\
		2st ATV   & 85.22$\%$ & 3.83mm  & \makecell{8} & 0.48mm\\
		\Xhline{2pt}
	\end{tabular}
	\caption{Evaluation of uncertainty estimation. 
	The PSV is the first-stage plane sweep volume; 
	the 1st ATV is constructed after the first stage and used in the second stage; the 2nd ATV is used in the third stage.
	We show the percentages of uncertainty intervals that cover the ground truth depth. 
	We also show the average length of the intervals, the number of depth planes
	and the unit sampling distance.
	}
	\label{tab:uncertainty}
\end{table}

We can see that our method is able to construct efficient ATVs that cover very local depth ranges.
The first ATV significantly reduces the initial depth range from 508.8mm to only 13.88mm in average, 
and the second ATV further reduces it to only 3.83mm.
Our ATV enables efficient depth sampling in an adaptive way, 
and obtains about 0.48mm sampling distance with only 32 or 8 depth planes.
Note that, MVSNet and R-MVSNet sample the same large depth range (508.8mm) in a uniform way with a large number of planes (256 and 512);
yet, the uniform sampling merely obtains volumes with sampling distances of 1.99mm and 0.99mm along depth.
In contrast, our UCS-Net achieves a higher actual depth-wise sampling rate with a small number of planes; 
this allows for the focus of the cost volumes to be changed 
from sampling the depth to sampling the image plane with dense pixels in ATVs given the limited memory, 
which enables high-resolution depth reconstruction.

Besides, 
our adaptive thin volumes achieve high ratios ($94.72\%$ and $85.22\%$) of 
covering the ground truth depth in the validate set, as shown in Tab.~\ref{tab:uncertainty};
this justifies that our estimated uncertainty intervals are of high confidence.
Our variance-based uncertainty estimation is equivalent to 
approximating a depth probability distribution as a Gaussian distribution and then
computing its confidence interval with a specified scale on its standard deviation as in Eqn.~\ref{eqn:interval}.

We note that our variance-based uncertainty estimation is not only valid for single-mode Gaussian-like distributions as in Fig.~\ref{fig:method}.a, 
but also valid for many multi-mode cases as in Fig.~\ref{fig:method}.b, which shows a challenging example near object boundary.
In Fig.~\ref{fig:method}.b, the predicted first-stage depth distribution has multiple modes; 
yet, it correspondingly has large variance and a large enough uncertainty interval.
Our network predicts reasonable uncertainty intervals 
that are able to cover the ground truth depth in most cases, 
which allows for increasingly accurate reconstruction in the following stages at finer local spatial scales.
This is made possible by the differentiable uncertainty estimation and the end-to-end training process,
from which the network learns to control per-stage probability estimation to obtain proper uncertainty intervals for ATV construction.
Because of this, we observe that our network is not very sensitive to different $\lambda$, and learns to predict similar uncertainty.
Our uncertainty-aware volume construction process enables highly efficient spatial partitioning, 
which further allows for the final reconstruction to be of high accuracy and high completeness.

\begin{table}[t!]
	\footnotesize
	\centering
	\begin{tabular}[width=\linewidth]{p{0.5cm}p{0.5cm}p{1.5cm}|p{1.05cm}p{1.05cm}p{1.2cm}}
		\Xhline{2pt}
		Stage & Scale & Size & Acc. & Comp. & Overall\\ 
		\Xhline{2pt}
        1 & $\times$1 & 400x296& 0.548 & 0.529 & 0.539\\
		1 & $\times$2 & 800x592&  0.411 & 0.535 & 0.473 \\
        2 & $\times$1 & 800x592   & 0.401 & 0.397 & 0.399\\
		2 & $\times$2 & 1600x1184 & 0.342  & 0.386 & 0.364 \\
        3 & $\times$1 & 1600x1184 & 0.338 & 0.349 & 0.344\\
		\Xhline{2pt}
	\end{tabular}
	\caption{Ablation study on the DTU testing set with different stages and upsampling scales (a scale of 1 represents the original result at the stage). 
	The quantitative results represent average distances in mm (lower is better). }
	\label{tab:stage}
\end{table}

\begin{figure*}[t]
	\centering

    \includegraphics[width=\linewidth]{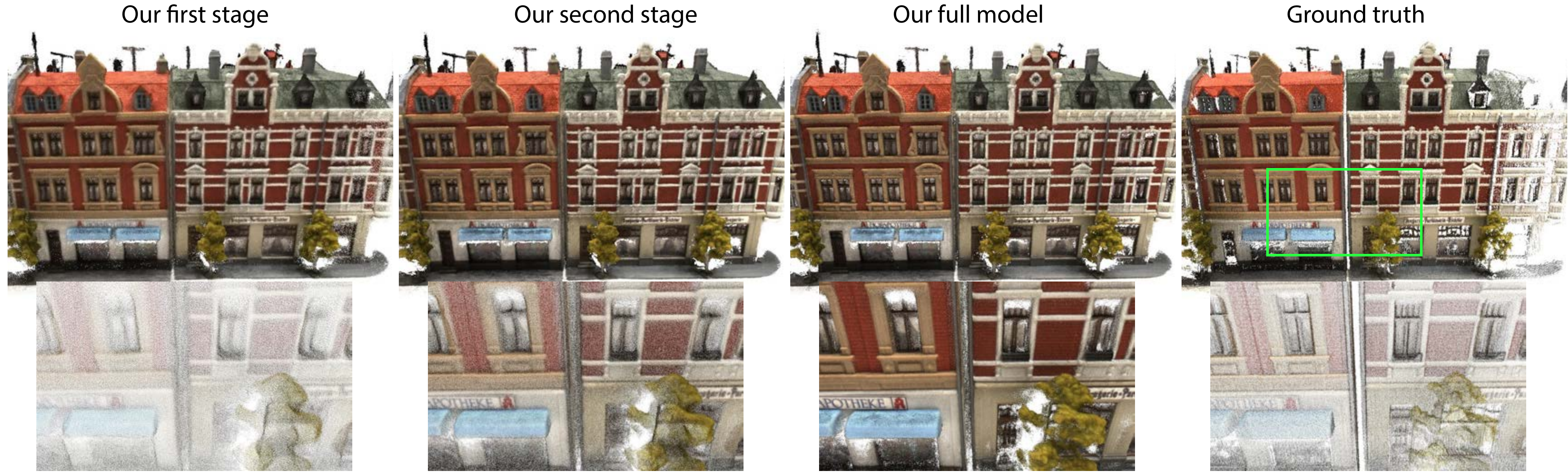}
	\caption{
		Qualitative comparisons between multi-stage point clouds and the ground truth point cloud on a scene in the DTU validate set. 
		We show zoom-out (top) and zoom-in (bottom) rendered point clouds; the corresponding zoom-in region is marked in the ground truth as a green box.
		Our UCS-Net achieves increasingly dense and accurate reconstruction through the multiple stages. Note that, the ground truth point cloud is obtained by scanning, which is even of lower quality than our reconstructions in this example.
}
	\label{fig:multistage}
\end{figure*}

\boldstartspace{Evaluation of multi-stage depth prediction.}
We have quantitatively demonstrated that our multi-stage framework reconstructs scene geometry with increasing accuracy and completeness in every stage (see Fig.~\ref{tab:dtu}).
We now further evaluate our network and do ablation studies about different stages on the DTU testing set with detailed quantitative and qualitative comparisons.
We compare with naive upsampling to justify the effectiveness of our uncertainty-aware coarse-to-fine framework. 
In particular, we compare the results from our full model and the results from the first two stages with naive bilinear upsampling using a scale of 2 (for both height and width) in Tab.~\ref{tab:stage}.
We can see that upsampling does improve the reconstruction, which benefits from denser geometry and using our high-quality low-resolution results as input.
However, the improvement made by naive upsampling is very limited, which is much lower than our improvement from our ATV-based upsampling.
Our UCS-Net makes use of the ATV -- a learned local spatial representation that is constructed in an uncertainty-aware way 
-- to reasonably densify the map with a
significant increase of both completeness and accuracy at the same time.

Figure.~\ref{fig:multistage} shows qualitative comparisons between our reconstructed point clouds and the ground truth point cloud.
Our UCS-Net is able to effectively refine and densify the reconstruction through multiple stages.
Note that, our MVS-based reconstruction is even more complete than the ground truth
point cloud that is obtained by scanning, which shows the high quality of our reconstruction.

\begin{table}[t!]
	\footnotesize
	\centering
	\begin{tabular}[width=\linewidth]{c|p{1.05cm}p{1.05cm}p{1.1cm}p{1.1cm}}
		\Xhline{2pt}
		Method & Running time (s) & Memory (MB) & Input size & Prediction size\\ 
		\Xhline{2pt}

		\makecell{One stage \\ 
				  Two stages \\ 
				  Our full model}  &\makecell{ 0.065 \\ 
				                                0.114 \\ 
											    0.257 } &\makecell{ 1309 \\
											                         1607 \\
											                         1647 } & 640x480 &                                     \makecell{160x120 \\
																					320x240 \\
																					640x480} \\
		\Xhline{2pt}
        MVSNet \cite{yao2018mvsnet} &   \makecell{1.049} &     \makecell{4511} & 640x480 & 160x120\\

		R-MVSNet \cite{yao2019recurrent} &  \makecell{1.421} &     \makecell{4261} & 640x480 & 160x120\\
		
		\Xhline{2pt}
	\end{tabular}
	\caption{Performance comparisons. We show the running time and memory of our method by running the first stage, the first two stages and our full model.}
	\vspace{-2mm}
	\label{tab:time}
\end{table}

\boldstartspace{Comparing runtime performance.}
We now evaluate the timing and memory usage of our method.
We run our model on the DTU validate set with an input image resolution of $W\times H=640\times 480$;
We compare performance with MVSNet and R-MVSNet with 256 depth planes using the same inputs.
Table~\ref{tab:time} shows the performance comparisons including running time and memory.
Note that, 
our full model is the only one that reconstructs the depth at the original image resolution that is much higher than the comparison methods.  
However, this hasn't introduced any higher computation or memory consumption. 
In fact, our method requires significantly less memory and shorter running time, which are only about a quarter of the memory and time used in other methods.
This demonstrates the benefits of our coarse-to-fine framework with fewer depth planes (104 in total), in terms of system resource usage.
\comment{
which is significantly lower than MVSNet and R-MVSNet.
Because of the usage of significantly 
While our full model requires longer running time than the other two methods, our first two stages alone perform with timing comparable to others.
Note that, our two-stage reconstruction has already achieved better reconstruction than
the comparison methods as shown in Tab.~\ref{tab:dtu}.
}
Our UCS-Net with ATVs achieves high-quality reconstruction with high computation and memory efficiency.

\section{Conclusion}
In this paper, we present a novel deep learning-based
approach for multi-view stereo. 
We propose the novel uncertainty-aware cascaded stereo network (UCS-Net), which utilizes the adaptive thin volume (ATV), a novel spatial representation.
For the first time, we make use of the uncertainty of the prediction in a learning-based MVS system. 
Specifically, we leverage variance-based uncertainty intervals at one cascade stage to construct an ATV for its following stage.
The ATVs are able to progressively sub-partition the local space at a finer scale, and ensure that the smaller volume still surrounds the actual surface with a high probability. 
Our novel UCS-Net achieves highly accurate and highly complete scene reconstruction in a coarse-to-fine fashion.
We compare our method with various state-of-the-art benchmarks; we demonstrate that our method is able to achieve the qualitatively and quantitatively best performance with moderate computation- and memory- complexity.
Our novel UCS-Net takes a step towards making the learning-based MVS method more reliable and efficient.

\boldstartspace{Acknowledgements}
This work was funded in part by Kuaishou Technology, NSF grant IIS-1764078, NSF grant 1703957, the
Ronald L. Graham chair and the UC San Diego Center for Visual Computing.

{\begin{flushleft}\LARGE \textbf{Appendix} \end{flushleft}}

\section*{Overview}
In this appendix, we evaluate the uncertainty estimation with additional experiments, show the sub-networks of our network architecture in detail, and
demonstrate our final point cloud reconstruction results of the DTU testing set and the Tanks and Temple dataset. 

\section{Additional experiments of uncertainty estimation.}
In this section, we discuss additional experiments and analysis about our uncertainty estimation evaluated on the DTU validate set.

\begin{figure}[h]
    \centering
    \includegraphics[width=\linewidth]{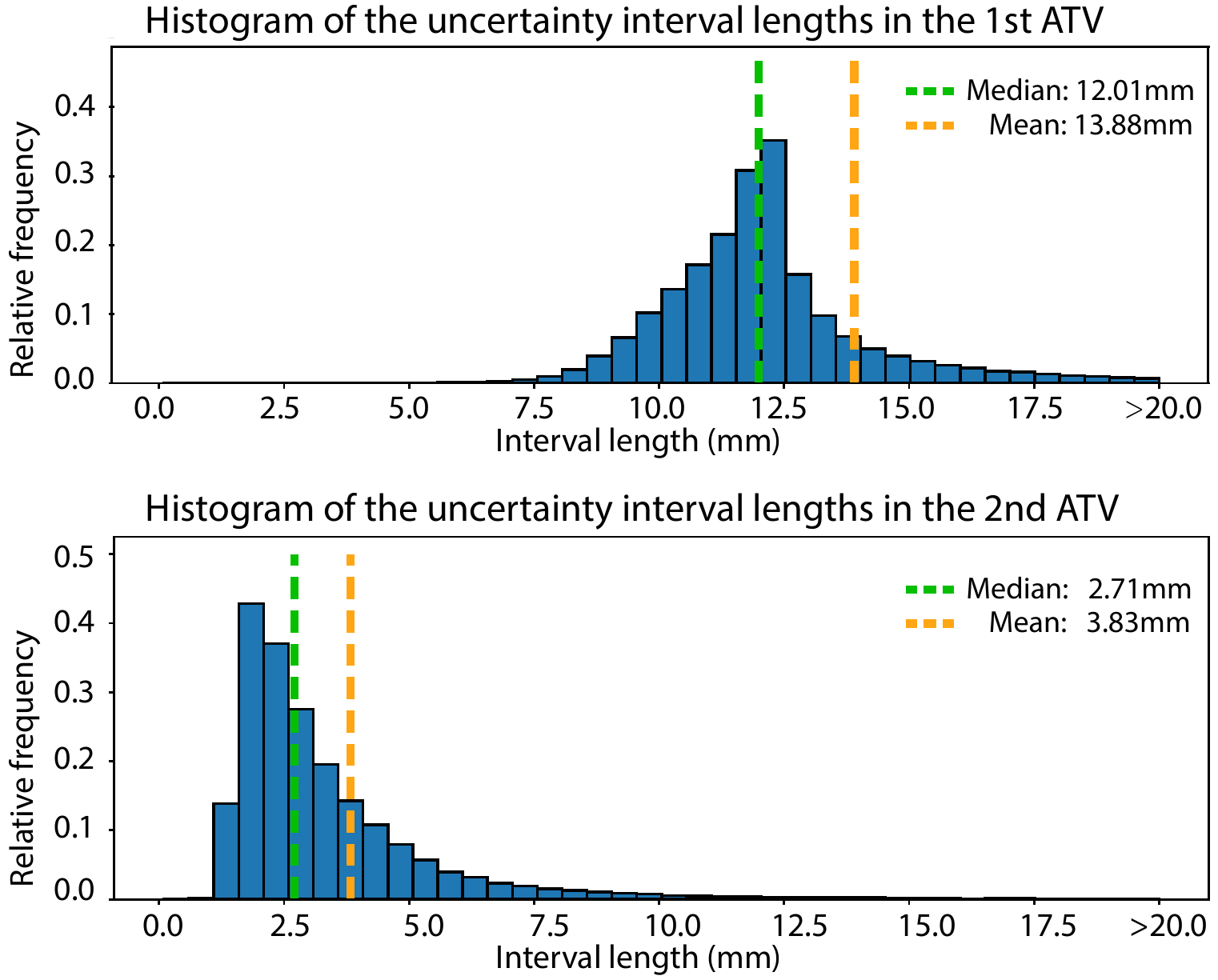}
    \caption{Histograms of the uncertainty interval lengths. 
    We create bins for every 0.5mm to compute the histograms of the lengths of the uncertainty intervals in the two ATVs.
    We mark the median and the mean values of the lengths in the histograms.
	}
    \label{fig:hist}
\end{figure}

\begin{table}[h!]
	\centering
	\includegraphics[width=\linewidth]{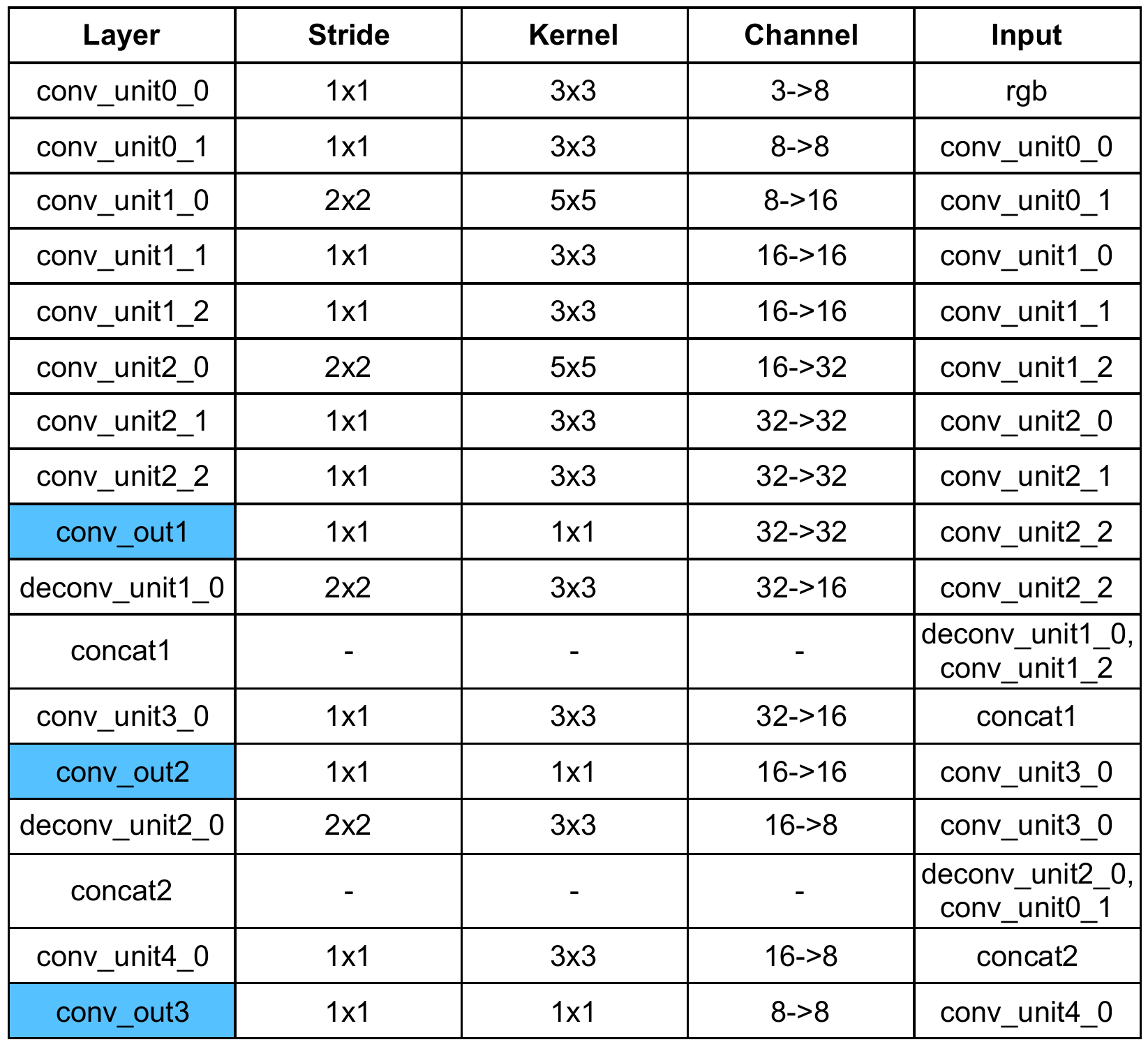}
	\caption{
		The U-Net architecture of our multi-scale feature extractor. We show the detailed convolutional units of our multi-scale feature extractor; each convolutional unit is composed by a 2D convolution layer, a BN (batch normalization) layer and a ReLU layer. The colored cells (conv\_out1, conv\_out2, conv\_out3) apply only a single 2D convolution layer to provide multi-scale features for cost volume construction.
		}
	\label{fig:2d_unet}
\end{table}

\begin{table}[h!]
	\centering
	\includegraphics[width=\linewidth]{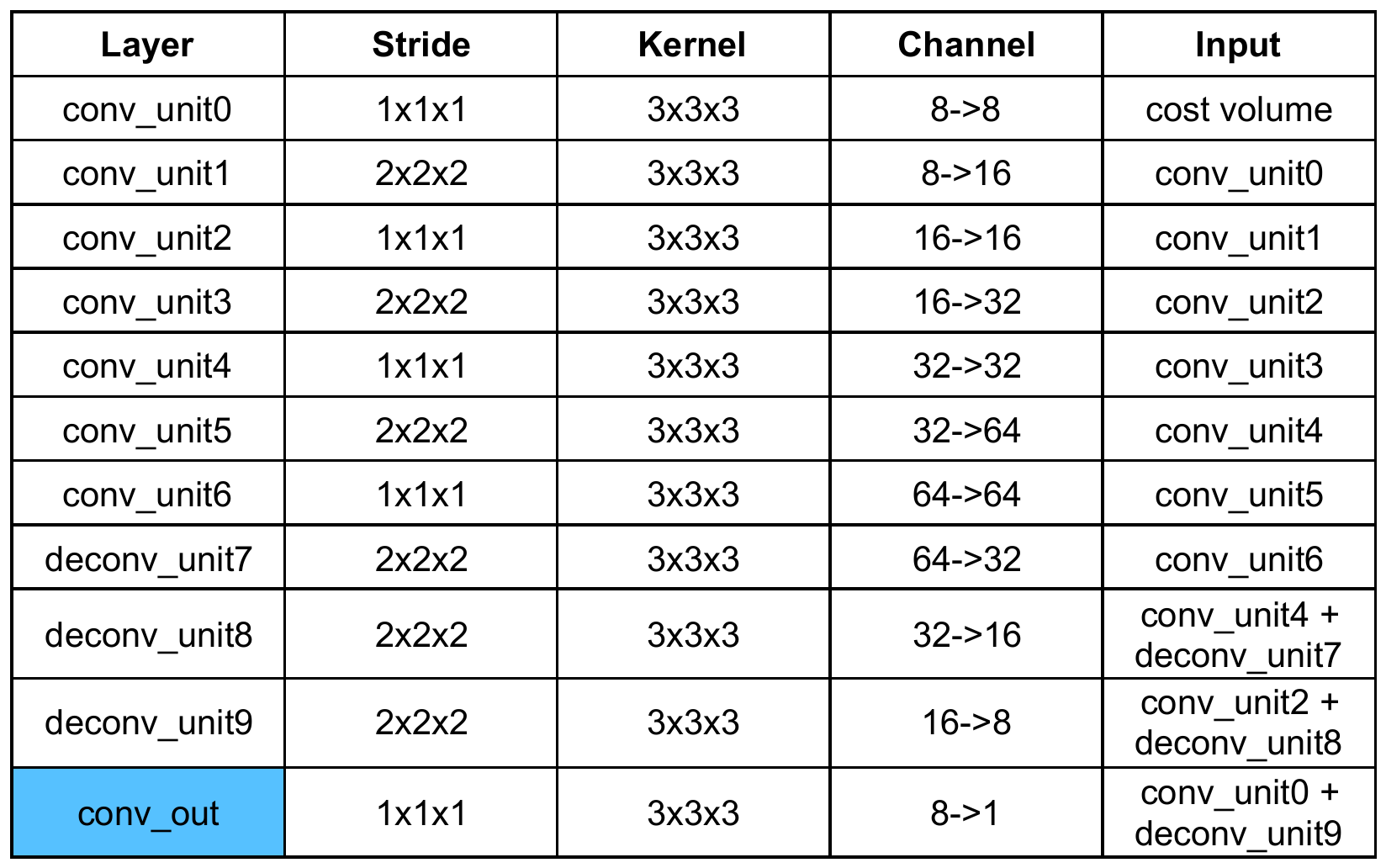}
	\caption{
		The network architecture of the 3D U-Net. We show the 3D U-Net architecture that is used to process the cost volume and predict the depth probabilities at each stage. Similarly, each convolutional unit is composed by a 3D convolution layer, a BN (batch normalization) layer and a ReLU layer. The colored cell (conv\_out) apply only a single 3D convolution layer. We apply soft-max on the final one-channel output over depth planes to compute the final depth probability maps. 
		}
	\label{fig:3d_unet}
\end{table}

\begin{figure*}[t!]
    \centering
    \includegraphics[width=\linewidth]{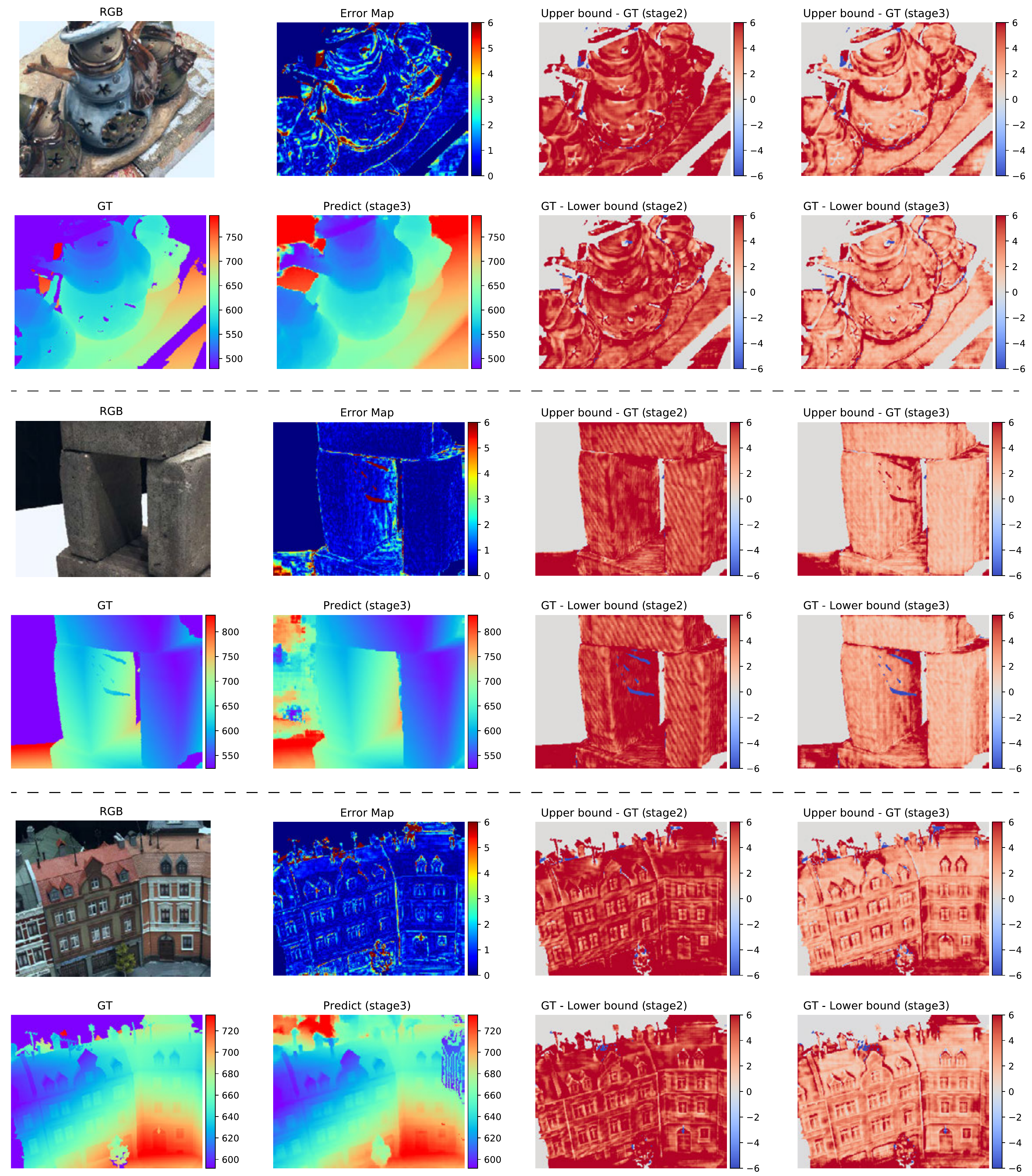}
    \caption{Uncertainty in depth predictions. We show three examples from the DTU validate set regarding the depth predictions and their pixel-wise uncertainty estimates.
    In each example, we show the reference RGB image, the ground truth depth, the depth prediction and a corresponding error map;
    we also illustrate the uncertainty intervals by showing the difference between the ground truth depth and the interval boundaries (lower bound and upper bound).
    Note that, in the right two columns, the white colors represent small intervals with low uncertainty,
    the red colors represent large intervals with large uncertainty, 
    and the blue colors correspond to the intervals that fail to cover the ground truth.   
	}
    \label{fig:bound}
\end{figure*}
We have shown the average lengths of the uncertainty intervals and
the corresponding average sampling distances between the depth planes of the ATVs in Tab.~3 of the main paper.
We now show the histograms of the uncertainty interval length in Fig.~\ref{fig:hist} to better illustrate the distributions of the interval length. 
We also mark the average lengths and the median lengths in the histograms.
Note that, the distributions of the two ATVs are unimodal, in which most lengths distribute around the peaks; however, 
the average interval lengths differ much from the modes in the histograms,
because of small portions of the intervals that have very large uncertainty.
This means that using the average interval lengths -- as what we do for Tab.~3 in the main paper -- to discuss the depth-wise sampling is in fact underestimating the sampling efficiency we have achieved for most of the pixels, though our average lengths are good and correspond to a high sampling rate.
Therefore, we additionally show the median values in the histograms, which are less sensitive to the large-value outliers and
are more representative than the mean values for these distributions.
As shown in Fig.~\ref{fig:hist}, the median interval lengths of the two ATVs are 12.01mm and 2.71mm respectively, which are closer to the peaks of the histograms;
these lengths correspond to depth-wise sampling distances of 0.38mm and 0.34mm, given our specified 32 and 8 depth planes.
These are significantly higher sampling rates than previous works, 
such as MVSNet \cite{yao2018mvsnet} -- which uses 256 planes to obtain a sampling distance of 1.99mm -- and RMVSNet \cite{yao2019recurrent} -- which uses 512 planes to obtain a sampling distance of 0.99mm.
Our ATV allows for highly efficient spatial partitioning, which achieves a high sampling rate with a small number of depth planes.

To illustrate how the per-pixel uncertainty estimates vary in a predicted depth map,
we show the pixel-wise difference between the ground truth depth and the boundaries of the uncertainty intervals in Fig.~\ref{fig:bound}.
We can see that, while our estimated uncertainty intervals have small lengths (as shown in Fig.~\ref{fig:hist}),
the uncertainty estimation is very reliable, reflected by the fact that most intervals are covering the ground truth depth in both ATVs (the red and white colors in the right two columns of Fig.~\ref{fig:bound}) .
This verifies the high average covering ratios of $94.7\%$ and $85.2\%$ of the two ATVs, which we have shown in Tab.~3 of the main paper.
We also observe more white colors in the 3rd-stage ATV than those in the 2nd-stage ATV, which reflects that the uncertainty is well reduced after a stage and the prediction becomes more precise.
Note that, while our method may predict incorrect intervals (blue colors in the right two columns of Fig.~\ref{fig:bound}) that fail to cover the ground truth for some pixels, those pixels are mostly around the shape boundaries, oblique surfaces and highly textureless regions, which are known to be challenging and still open problems for depth estimation.
On the other hand, our method predicts large uncertainty for these challenging pixels, which is as we expect and reflects the inaccuracies in the predictions.


\section{Network architecture.}



We have shown the overview of our network in Fig.~2 of the main paper and discussed our network in Sec.~3 of the paper. 
Our network consists of a 2D U-Net for feature extraction and three 3D U-Nets with the same architecture for cost volume processing. 
We show the details of the 2D U-Net in Tab.~\ref{fig:2d_unet}, which is used for our multi-scale feature extractor (see Sec.~3.1 of the paper); we also show the details of our 3D U-Net in Tab.~\ref{fig:3d_unet} which is used to process the cost volume at each stage (see Sec.~3.3 of the paper).

\section{Point cloud reconstruction.}
\begin{figure*}[t]
    \centering
    \includegraphics[width=2\columnwidth]{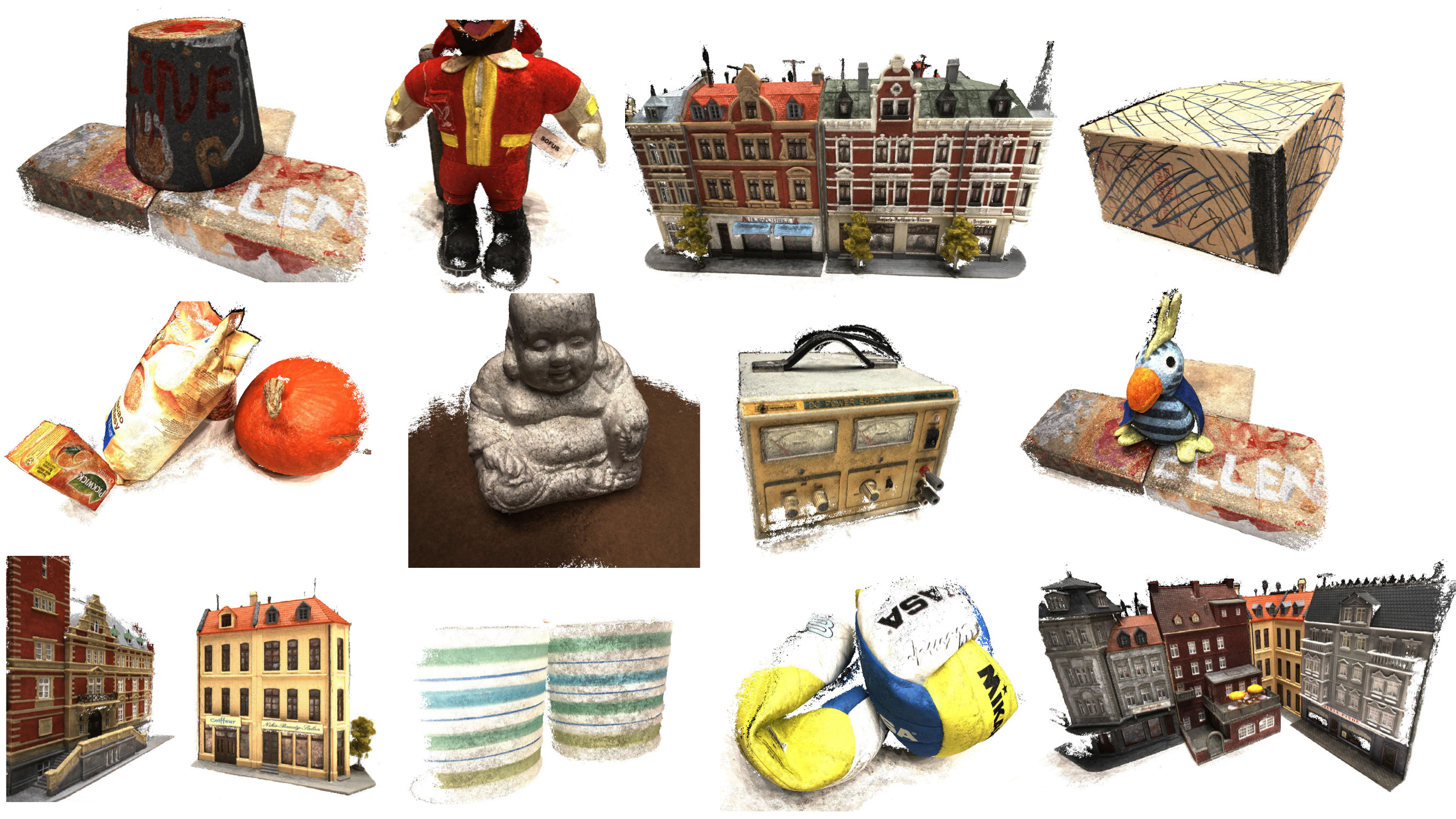}
    \caption{Point cloud reconstruction on the DTU test set.}
    \label{fig:dtu1}
\end{figure*}

\begin{figure*}[t]
    \centering
    \includegraphics[width=2\columnwidth]{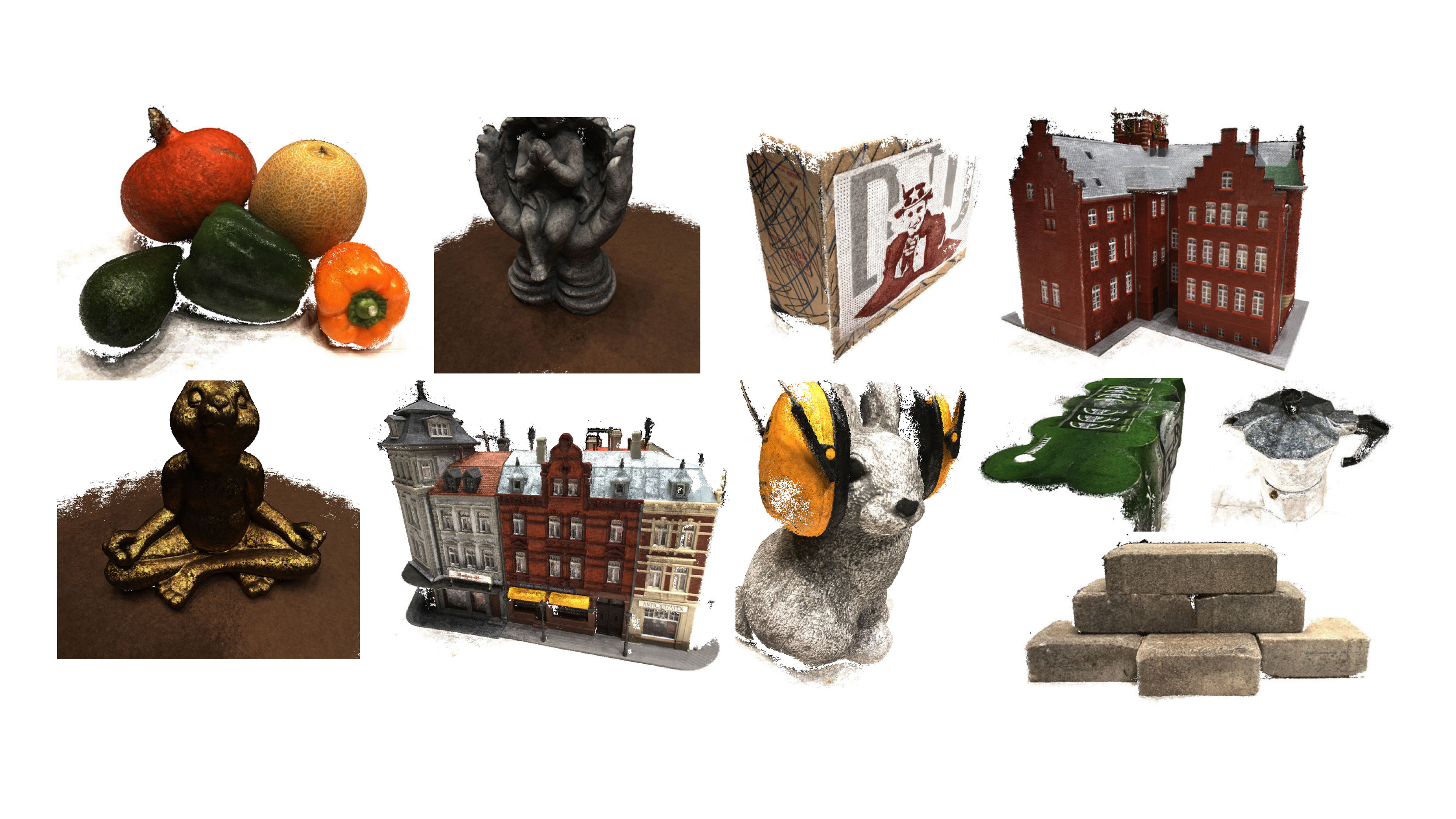}
    \caption{Point cloud reconstruction on the DTU test set.}
    \label{fig:dtu2}
\end{figure*}

\begin{figure*}[t]
    \centering
    \includegraphics[width=2\columnwidth]{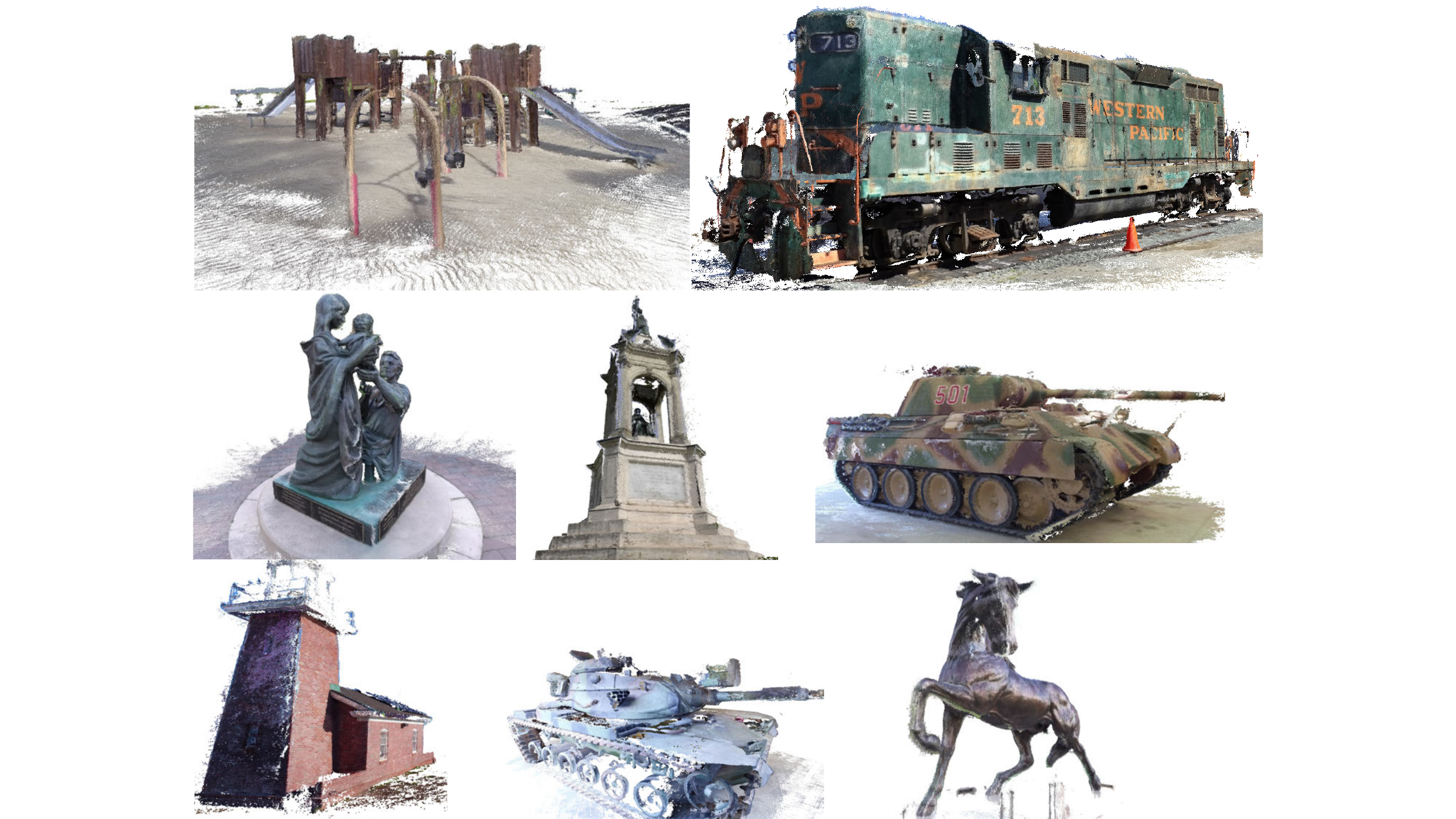}
    \caption{Point cloud reconstruction on the Tanks and Temple dataset.}
    \label{fig:tt}
\end{figure*}

We show our final point cloud reconstruction results of the DTU testing set \cite{aanaes2016large} in Fig.~\ref{fig:dtu1} and Fig.~\ref{fig:dtu2}, and the results of the Tanks and Temple dataset \cite{knapitsch2017tanks} in Fig.~\ref{fig:tt}.
Please refer to Tab.~1 and Tab.~2 in the main paper for quantitative results on these datasets.


{\small
\bibliographystyle{ieee_fullname}
\bibliography{egbib}
}

\end{document}